\documentclass[11pt]{article}

\usepackage[preprint]{acl}

\usepackage{times}
\usepackage{latexsym}

\usepackage[T1]{fontenc}

\usepackage[utf8]{inputenc}

\usepackage{microtype}

\usepackage{inconsolata}

\usepackage{graphicx}

\usepackage{amsmath}
\usepackage{amsfonts}
\usepackage{booktabs}
\usepackage{multirow}
\usepackage{hyperref}
\usepackage[skins, breakable]{tcolorbox}

\usepackage{amsthm}
\newtheorem{proposition}{Proposition}
\newtcolorbox{prbox}[1][]{
    enhanced,
    after skip=8mm, 
    title=#1,
    breakable = true,
    fonttitle=\sffamily\bfseries\color{white},
    coltitle=white,
    colbacktitle=gray!80!white, 
    titlerule=0pt,
    overlay={%
        \ifcase\tcbsegmentstate
        \or%
        \else%
        \fi%
    },
    colback = gray!2!white,          
    colframe = gray!80              
}
\usepackage{xcolor}
\newcommand{\algo}{ARR}

\title{Adversarial Yet Cooperative: Multi-Perspective Reasoning in Retrieved-Augmented Language Models}

\author{
 \textbf{Can Xu\textsuperscript{1,2}},
 \textbf{Lingyong Yan\textsuperscript{2}},
 \textbf{Jiayi Wu\textsuperscript{1}},
 \textbf{Haosen Wang\textsuperscript{3}},
 \textbf{Shuaiqiang Wang\textsuperscript{2}},
\\
 \textbf{Yuchen Li\textsuperscript{2}},
 \textbf{Jizhou Huang\textsuperscript{2}},
 \textbf{Dawei Yin \textsuperscript{2}},
 \textbf{Xiang Li\textsuperscript{1}},
\\
\\
 \textsuperscript{1}East China Normal University,
 \textsuperscript{2}Baidu Inc.,
 \textsuperscript{3}Southeast University
\\
 \small{
   \textbf{Correspondence:} Xiang Li \ \href{xiangli@dase.ecnu.edu.cn}{xiangli@dase.ecnu.edu.cn}
 }
}

\begin{document}
\maketitle
\begin{abstract}
      Recent advances in synergizing large reasoning models (LRMs) with retrieval-augmented generation (RAG) have shown promising results, yet two critical challenges remain: (1) reasoning models typically operate from a single, unchallenged perspective, limiting their ability to conduct deep, self-correcting reasoning over external documents, and (2) existing training paradigms rely excessively on outcome-oriented rewards, which provide insufficient signal for shaping the complex, multi-step reasoning process. To address these issues, we propose an Reasoner-Verifier framework named Adversarial Reasoning RAG (\algo). The Reasoner and Verifier engage in reasoning on retrieved evidence and critiquing each other's logic while being guided by process-aware advantage that requires no external scoring model. This reward combines explicit observational signals with internal model uncertainty to jointly optimize reasoning fidelity and verification rigor. Experiments on multiple benchmarks demonstrate the effectiveness of our method. Our code is available at \href{https://github.com/LEOXC1571/Code-of-ARR}{link}.
\end{abstract}

\section{Introduction}
\label{sec:introduction}
Large language models (LLMs) endowed with step-by-step reasoning capabilities have achieved remarkable success in complex question answering, especially when augmented with external knowledge through retrieval-augmented generation (RAG)~\citep{webthinker_25_li2,searcho1_25_li,retool_25_feng,ragen_25_wang}. Different from previous RAG methods that focus on retrieval optimization and component-based architectural design, recent efforts have been made on post-training LLM agents \citep{searchr1_25_jin,coa_25_li} integrated with search tools. 

Despite the effectiveness,
current RAG mainly adopts a monologic reasoning architecture, where only one single LLM-based agent reasons and interacts with search engines. However, when retrieved documents are partial, conflicting or misleading, the single-view reasoning may amplify errors rather than mitigate them. Prior efforts address this challenge by incorporating self-verification process~\citep{webseer_25_he,researcher_25_fu}. However, such self-critique paradigm also suffers from the single-view architecture, as many studies~\citep{llmjudge_24_xu,llmjudge_25_wu,llmjudge_25_zhang} show that LLMs struggle to identify their own logical flaws.


Moreover, in order to train the agentic RAG system, most existing methods optimize the RL framework using outcome-oriented, task-level rewards (e.g., accuracy or format correctness). Such rewards assign uniform reward to tokens within a sequence based on the final correctness, lacking supervision for the intermediate process. 
Unlike self-contained trajectories in mathematical domains,
the correctness in RAG system depends not only on reasoning quality, but on external factors beyond the agent's control, such as the precision of retrieval engine, the consistency of external documents, and the presence of conflicting evidence. 
Therefore, outcome-based rewards cannot distinguish between a correct answer derived through sound logic and the one produced by lucky guesswork, nor can they penalize plausible but flawed reasoning that happens to yield a wrong answer.

To tackle these challenges, we propose \textbf{\algo} (\textbf{A}dversarial \textbf{R}easoning \textbf{R}AG), a multi-perspective framework that {explicitly decouples reasoning and verification into separate perspectives}, handled by a reasoner agent and a verifier agent, respectively.
And we formalize such interactive process as an adversarial yet collaborative dialogue between them:

\textbf{Adversarial yet cooperative interaction}: The two agents should challenge each other not for winning the debate, but for a shared objective. Critiques should be justified and evidence-grounded.

\textbf{Process-aware learning}: The two agents are rewarded not only for correct final answers, but also for high-quality interactive process between them (e.g., logical coherence, evidence utilization, and uncertainty reduction).

To this end, we introduce an \emph{adversarial outcome reward} and a \emph{process-aware advantage} into the co-evolving process of both agents. 
(1). The adversarial outcome reward encourages agents to compete for higher correctness, ensuring that the consensus is driven by rigorous debate rather than blind agreement.
(2). The process-aware advantage is a token-level advantage for the verifier, which is driven by a core insight: high-quality reasoning in RAG should mirror the reduction of uncertainty and semantic entropy. With the proposal of search queries and the accumulation of evidences, the agent moves from an initial state of confusion to a state of crystallization. 
Based on this guiding principle, the process-aware advantage assesses the soundness of response, the clarity of verification, and the impact on the reasoner's cognitive state. By monitoring the evolution of reasoner policy entropy, we reward verifier's feedback that is confident, evidence-grounded, and steers the reasoner from high-entropy exploration to low-entropy convergence, thereby aligning the optimization with the information gain.

In summary, our main contributions are:

(1) We propose \algo, where reasoner and verifier engage in adversarial yet cooperative dialogue.

(2) We propose adversarial outcome reward to promote rigorous debate between agents, which encourages agents to compete for higher correctness.

(3) We design the token-level process-aware advantage. By modeling reasoning progress as the reduction of uncertainty, we reward trustworthy and evidence-grounded verifier feedback that effectively steers the reasoner toward better reasoning.

\section{Related Work}
\label{sec:related_works}

\subsection{Reward Design and Process Supervision}
\label{subsec:rewarddesign}
Reinforcement Learning with Verifiable Rewards (RLVR) has emerged as a powerful approach to enhance the reasoning capabilities of LLMs.
For example,
Pass@k~\citep{passk_25_chen} reveals that outcome rewards provides limited learning signals for tasks that are either overly simple or difficult, and fail to discriminate between effective and ineffective process within the reasoning trace. It leverages pass@k performance as the replacement for outcome only rewards. DAPO~\citep{dapo_25_yu} introduces dynamic sampling to filter out samples where model consistently succeeds or fails. 
In reasoning RAG scenrario, Atom-Searcher~\citep{atomsearch_25_deng} introduces reasoning reward model to provide process signal additional to outcome reward. WebSeer~\citep{webseer_25_he} introduces F1-score as the intermediate-step verification signal to guide the exploration process of search agent. 

\subsection{LRMs Synergizied with RAG}
\label{subsec:lrmrag}
Recent advances in RAG systems include the integration of search tools and LRMs, which significantly improve the capabilities for complex and multi-step reasoning and searching. 
The representative methods
Search-R1 \citep{searchr1_25_jin} and R1-Searcher \citep{r1searcher_25_song} train models to automatically derive reasoning through multi-turn searching. DeepResearcher~\citep{deepre_25_zheng} further includes web search agent into agentic reasoning RAG. Existing methods are primarily built upon single-agent frameworks, leaving a gap in exploration from multi-perspective interactions.

\section{Preliminary}
\label{sec:preliminary}

\subsection{Task Formulation}
\label{subsec:task_formulation}
An ideal agentic reasoning RAG system should go beyond the search-retrieve-answer pipeline and possess high-order capabilities, including: 

\textbf{Critical reasoning}: the capability to assess the reliability of external evidence and detect logical flaws in reasoning traces;

\textbf{Grounded generation}: the ability to anchor reasoning in verifiable evidence, and to revise conclusions when support is insufficient;

\textbf{Iterative refinement}: the ability to enhance reasoning quality through self- and peer-assessment, balancing both accuracy and process behaviors.

Current agentic RAG systems, however, remain constrained by monologic architectures and optimization objectives that rely predominantly on scaler outcome rewards. 
To bridge this gap, We propose \algo, a multi-perspective reasoning framework in which two agents learn to reason not as single voices, but through interaction of different viewpoints.
Formally, we model the system as a multi-agent Markov Decision Process (MDP), defined as the tuple $(\mathcal{S}^\alpha$, $\mathcal{A}^\alpha$, $\mathcal{P}^\alpha$, $\mathcal{R}^\alpha)$. Let $\alpha \in \{\text{r,v}\}$ index the \textbf{Reasoner} ($\text{r}$) and the \textbf{Verifier} ($\text{v}$), interacting in an environment that includes search engine and document corpus $D$. Given a query $q$, the agent behavior is governed by the policy model $\pi_\theta^\alpha$. State $s_t^\alpha \in \mathcal{S}^\alpha$ refers to previous histories and external context by the agent other than $\alpha$, and $a_t^\alpha \in \mathcal{A}^\alpha$ is the action generated by $\pi_\theta^\alpha$ from its action space at turn $t$:
$a_t^\alpha = \pi_\theta^\alpha(s_t^\alpha)$.
Notably, distinct from token-level MDPs, we define the action space $\mathcal{A}^\alpha$ at the semantic step level. An action $a_t^\alpha$ is a sequence of tokens representing a complete move. Take Search-R1 for an example, the action space $\mathcal{A}^\alpha$ = \{\texttt{think}, \texttt{search}, \texttt{answer}\}.
A complete trace $\tau$ of $n$ interaction steps is denoted as
$\tau = ( s^{\text{r}}_1$, $a^{\text{r}}_1$, $s^{\text{v}}_1$, $a^{\text{v}}_1$, $...$, $s^{\text{r}}_n$, $a^{\text{r}}_{n}$, $s^{\text{v}}_n$, $a^{\text{v}}_n )$.

\begin{figure}[t]
  \includegraphics[width=\columnwidth]{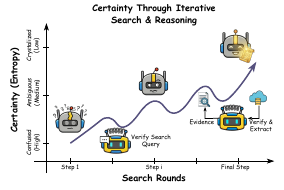}
  \caption{Ideal agent certainty through iterative search and reasoning}
  \label{fig:agentcertainty}
\end{figure}

\begin{figure*}[t]
\centering
  \includegraphics[width=1.0\linewidth]{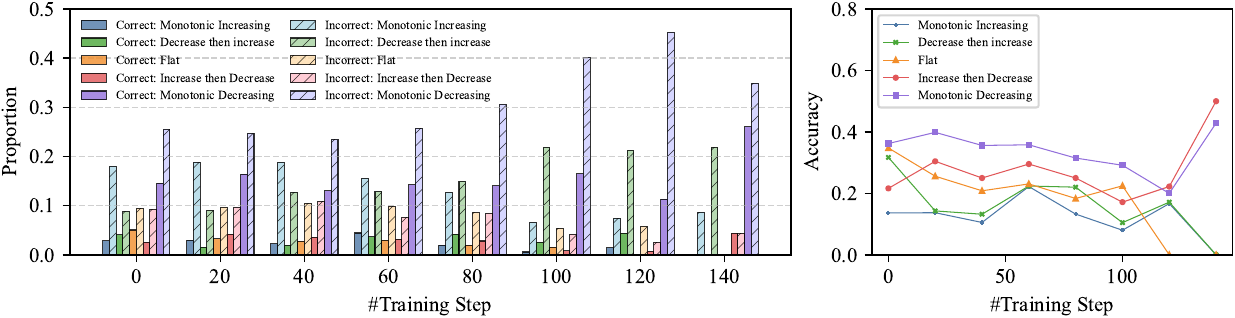}
  \caption {Statistical analysis of policy entropy pattern in Search-R1 trajectories. The y-axis of the \textbf{left subplots} denotes the proportion of trajectories exhibiting specific pattern in all multi-turn ($\ge 3$) samples. The y-axis of the \textbf{right subplots} represents the average accuracy of samples grouped by their pattens.}
  \label{fig:entpattern_acc}
\end{figure*}
\subsection{Entropy Pattern Analysis}
\label{subsec:ent_pattern}

Generally, RLVR for LLMs often involves the trade-off between policy entropy and performance \citep{entmech_25_cui}. In the context of RAG, the reasoning process can be regarded as the dynamic evolution of cognitive states driven by external knowledge management.
As illustrated in Figure~\ref{fig:agentcertainty}, an ideal reasoning trajectory exhibits three stages. (1) \emph{Initial uncertainty}: the agent begins in a confused state, where high policy entropy reflects the lack of knowledge and exploration of search queries. (2) \emph{Evidence integration}: the agent assimilates retrieval results and converges towards final answer. (3) \emph{Crystallization}: the agent has sufficient evidence and generates a well-supported conclusion.
To formalize this intuition, we present the following proposition.
\begin{proposition}
    In an ideal agentic RAG system, as relevant information is retrieved, both the uncertainty of agent and the policy entropy monotonically decrease.
\end{proposition}

\begin{proof}
Consider an agentic RAG system with single agent (e.g. Search-R1). Let $Y$ denotes the ground-truth answer. Given a user query $q$, the state $s_{t+1}$ is the union of prior state $s_t$, action $a_t$ and retrieved document $d_t$. To ensure rigor, we introduce two assumptions. 

\textbf{Assumption 1:} The agent acts to maximize the expected information gain and issues search queries intended to retrieve relevant documents. 

\textbf{Assumption 2:} The retrieved documents $d_t$ provides a positive information gain regarding $Y$.

\textit{(1). Monotonic Decrease of Answer Uncertainty:}
The uncertainty of $Y$ with document $d_t$ is quantified by the conditional mutual information:
\begin{equation}
    I(Y;d_t|s_{t},a_t) = H(Y|s_{t},a_t) - H(Y|s_{t+1}).
    \label{eq:mi_y_doc}
\end{equation}
By definition, the updated state is $s_{t+1} = [s_{t},a_t,d_t]$. Since $a_t$ is generated based on state $s_{t}$, we have the Markov property $H(Y|s_{t}, a_t) \approx H(Y|s_{t})$. Substituting these into Eq~\ref{eq:mi_y_doc}, we have:
$H(Y|s_{t+1}) \approx H(Y|s_{t}) - I(Y;d_t|s_{t},a_t).$
Since mutual information is non-negative, and the retriever provides relevant information, we obtain: 
\begin{equation}
    H(Y|s_{t+1}) \leq H(Y|s_{t}).
\end{equation}
This indicates that remaining uncertainty of the ground-truth is non-increasing with the accumulation of retrieved documents.

\textit{(2). Convergence of Policy Entropy:}
Next, we discuss the entropy of action $a_t$, noted as $H(a_t|s_{t})$. We decompose it using the definition of mutual information between action $a_t$ and ground-truth $Y$:
\begin{equation}
    H(a_t|s_{t}) = I(a_t;Y|s_{t}) + H(a_t|Y,s_{t}).
    \label{eq:ent_action}
\end{equation}
Then, we analyze the two terms on the right side of Eq~\ref{eq:ent_action}. First, the mutual information is defined as $I(a_t;Y|s_{t})= H(Y|s_{t}) - H(Y|a_t,s_{t})$. Thereby, $I(a_t;Y|s_{t}) \leq H(Y|s_{t})$ holds. Second, the term $H(a_t|Y,s_{t})$ represents the uncertainty of agent's action given that the ground truth is known. Following Assumption 2, as the training progresses, given the ground-truth answer, the agent would gradually come to a deterministic output,
i.e., $H(a_t| Y, s_{t}) \approx 0$. Therefore, we have the upper bound for the policy entropy:
\begin{equation}
    H(a_t|s_{t}) \leq H(Y|s_{t}) + H(a_t| Y, s_{t}).
\end{equation}
Since $H(Y|s_{t})$ is monotonically decreasing, and $H(a_t|s_{t})$ follows the upper bound of $H(Y|s_{t})$. This illustrates that as the agent accumulates evidence, its reasoning process naturally converges from exploration to exploitation.
\end{proof}

To empirically validate this theoretical proposition, we conduct a statistical analysis of the policy entropy evolution during the training process of Search-R1~\citep{searchr1_25_jin} in Figure~\ref{fig:entpattern_acc}\footnote{We only show results on Qwen2.5-3B here, and more results are shown in Appendix.}. Specifically, we focus on agent trajectories containing at least three search \& reasoning turns and aggregate the statistics every 20 training steps. The action entropy is $H_{a_t} = \frac{1}{|a_t|} \sum^{|a^\text{v}_t|}_{j=1} H(\pi_\theta (a_{t,j} | s_t, a_{t, < j}))$,
where $a_t$ $\in$ $\{\texttt{think}\}$ and $|\cdot|$ measures the sequence length of action $a_t$. Specifically, the trend between action $a_{t+1}$ and $a_t$ is \textit{Increase}, if $\Delta H_{a_{t+1}}$ > $\delta$; \textit{Decrease}, if $\Delta H_{a_{t+1}}$ < -$\delta$; \textit{Flat}, otherwise. Here, $\Delta H_{a_{t+1}} = H_{a_{t+1}} - H_{a_t}$ and $\delta$ is the threshold which accounts for minor fluctuations during reasoning. For each of them, we track the average token entropy of last three turns and categorize its evolution trend into five patterns: \textit{Monotonic Increasing} (I), \textit{Decrease-then-Increase} (DI), \textit{Flat} (F), \textit{Increase-then-Decrease} (ID), and \textit{Monotonic Decreasing} (D). We define a mapping function $f_e$: $R^n$ $\to$ $\{\text{D}$, $\text{ID}$, $\text{F}$, $\text{DI}$, $\text{I}\}$. Figure~\ref{fig:entpattern_acc} leads to two primary observations: 

\textbf{Correlation with Correctness:} There is a positive correlation between the \textit{Monotonic Decreasing} entropy pattern and model's accuracy. This suggests that effective reasoning is often accompanied by a progressive resolution of uncertainty.

\textbf{Evolution of Exploration:} Throughout training, there is a notable rise in the proportion of samples exhibiting an overall reduction in policy entropy (i.e., \textit{Increase-then-Decrease}, and \textit{Monotonic Decreasing}). Quantitatively, for the Qwen2.5-3B backbone, the proportion rises from 51.74\% in the early phase to 69.57\% in the late training phase. This suggests that the model learns to narrow down search space and converge on valid solutions as its multi-turn exploration capability deepens.

These observations support the premise that successful reasoning in RAG systems is intrinsically characterized by the progressive resolution of policy uncertainty. Collectively, this theoretical insight and empirical observation provide a robust foundation for the process reward design within our proposed multi-agent framework.
    

\begin{figure}[t]
\centering
  \includegraphics[width=1.0\columnwidth]{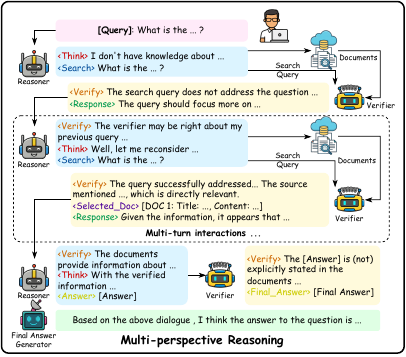}
  \caption {Multi-perspective reasoning of \algo.}
\end{figure}
\section{Methods}
\label{sec:methods}
\subsection{Multi-perspective Reasoning}
\label{subsec:mprea}
Building upon the formulation above, \algo\ performs multi-perspective reasoning and verification through an iterative dialogue between two agents:

\textbf{Reasoner} takes the lead in exploration. It formulates search queries, retrieves documents, constructs step-by-step reasoning, and proposes candidate answers. 

\textbf{Verifier} serves as a critical partner. It checks the relevance and credibility of search queries and retrieved documents, identifies logical gaps or unsupported claims in reasoning, and performs validation of the answers proposed by the reasoner. 

For the reasoner, the action space is defined as $\mathcal{A}^\text{r}$ = $\{ \texttt{think}$, $\texttt{search}$, $\texttt{verify}$, $\texttt{answer} \}$, a complete reasoning step is ($\texttt{think}$, $\texttt{search}$, $[\texttt{feedback}]$, $\texttt{verify}$), and the final step is ($\texttt{think}$, $\texttt{answer}$). 

$\bullet \ \texttt{think}$: the segment of reasoning grounded in the given query $q$ and retrieved evidence;

$\bullet \ \texttt{search}$: search queries issued when external knowledge is deemed necessary;

$\bullet \ [\texttt{feedback}]$: the feedback provided by the verifier, including supporting evidence or critiques;

$\bullet \ \texttt{verify}$: self-assessment of the reliability and sufficiency of external knowledge $\texttt{information}$.

$\bullet \ \texttt{answer}$: the answer when the reasoner thinks reasoning is complete and well-supported.
Correspondingly, the verifier operates within the action space $\mathcal{A}^\text{v}$ = $\{ \texttt{verify}$, $\texttt{selected\_doc}$, $\texttt{response}$, $\texttt{final\_answer} \}$. A complete verification step is ([$\texttt{information}$], $\texttt{verify}$, $\texttt{selected\_doc}$, $\texttt{response}$), and the final step is ($[\texttt{information}]$, $\texttt{verify}$, $\texttt{final\_answer}$).

$\bullet \ [\texttt{information}]$: search queries by the reasoner associated with retrieved documents or the reasoner's answer once it finishes reasoning;

$\bullet \ \texttt{verify}$: verification on validity of queries, document relevance, and logical soundness;

$\bullet \ \texttt{selected\_doc}$: curated documents (e.g. Doc n) that directly support or refute the claim;

$\bullet \ \texttt{response}$: explicit feedback to the reasoner, comprising either supporting evidence for valid queries or constructive critiques with justification for flawed ones.

$\bullet \ \texttt{final\_answer}$: the final conclusion after verifying all the evidence and the reasoner's answer.

Notably,  the reasoner has a built-in $\texttt{verify}$ stage within each step, allowing it to critically assess the reliability of retrieved evidence before coming to a conclusion.
The verifier is instructed to return the most relevant source passage in $\texttt{selected\_doc}$, rather than returning all retrieved documents to the reasoner without indiscriminately. This prevents information overload while guaranteeing feedback retains traceable and verifiable evidence.
Together, these mechanisms help form a balanced adversarial dialogue, where neither agent dominates, and reasoning quality emerges from their structured interactions. 

Following the iterative dialogue, we concatenate the full interaction history $\tau$ into a unified prompt, which is then fed to the final answer generator. Note that it employs the same policy model as the reasoner. By explicitly synthesizing insights from both perspectives, we obtain a more robust and well-grounded final answer. Detailed prompts for all agents are provided in Appendix~\ref{apd:prompt}.

\begin{figure}[t]
\centering
  \includegraphics[width=1.0\columnwidth]{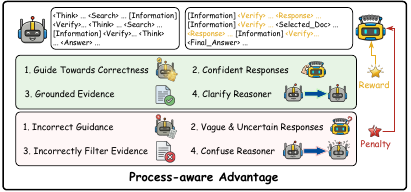}
  \caption {Process-aware advantage of \algo.}
\end{figure}

\subsection{Multi-perspective Optimization}
\label{subsec:mpoptim}
To overcome the limitations of sparse outcome-based supervision and to explicitly promote constructive adversarial interactions, we propose a multi-perspective reward design that disentangles final correctness from process fidelity. This design ensures that agents are rewarded not only for generating correct answers but also for engaging in high-quality and evidence-grounded dialogue. Our reward scheme consists of two components: adversarial outcome reward and process-aware advantage for the verifier. 

\paragraph{Adversarial Outcome Rewards}
In consistent with the adversarial yet cooperative dialogue design, our outcome reward explicitly promotes effective adversarial engagement by rewarding agents not just for its' correctness, but for outperforming their counterpart. Formally, each agent $\alpha$ receives an outcome reward composed of two terms:
\begin{equation}
    r^{\alpha} = \text{F1}(y^\alpha, y_{\text{gold}}) + \lambda \cdot \max \left[\mathtt{bin} (r^{\alpha}_{\text{out}} - r^{\bar{\alpha}}_{\text{out}}), 0\right],
\end{equation}
where $\bar{\alpha}$ denotes the counterpart agent, $y^\alpha$ denotes the answer by agent $\alpha$, and $y_{\text{gold}}$ is the ground truth. The operator $\mathtt{bin}(\cdot)$ discretizes the range of F1 score into $n$ buckets, filtering minor differences between answers by both agents. Thus, both agents are rewarded not only for correctness, but also for better performance than the other agent. Such reward also helps increase the discrimination of rewards within a group in Group Relative Policy Optimization (GRPO), particularly for tasks of moderate difficulty.

\paragraph{Process-aware Advantage}
While outcome reward drives both policy models towards accurate answers, they provide limited signal regarding how correctness is achieved. To address this, we introduce a token-level process-aware advantage $A^{\text{v}}_{proc}$ for the verifier, which encourages trustworthy and evidence-grounded response that steer the reasoner toward better reasoning. Formally, we define:
\begin{equation}
    A^{\text{v}}_{proc} = \text{F1}(y^\text{r}, y_{\text{gold}}) \cdot A_\text{clarity} \cdot A_\text{impact},
\end{equation}
where the three terms each encodes correctness, clarity, and behavior impact, respectively.

\subparagraph{(1) Answer Correctness}
The reasoner's final F1 score serves as a necessary condition: only when the dialogue yields a correct answer does the verifier receive full process credit. This prevents rewarding the verifier for critiques that lead reasoning toward wrong conclusions.

\subparagraph{(2) Verifier Clarity} 
\begin{align}
    A_\text{clarity} =  & \text{exp}(-H^\text{v}_{a_t}) \cdot \mathbb{I}[y_{\text{gold}} \ \text{in} \ d_t] \nonumber \\
    & \cdot (2 \mathbb{I}[y_{\text{gold}} \ \text{in} \ a^\text{v}_t] -1),
\end{align}
where $d_t$ $\in$ $D$ is the retrieved documents at step $t$, and $a^{\text{v}}_t$ is the verifier's action in $\mathcal{A}^\text{v}_\text{sub}$ (e.g., \{\texttt{verify}, \texttt{response}\}). Here, $H^\text{v}_{a_t}$ denotes the average policy entropy of action $a^{\text{v}}_t$:
$H^\text{v}_{a_t} = \frac{1}{|a^\text{v}_t|} \sum^{|a^\text{v}_t|}_{j=1} H(\pi^\text{v}_\theta (a^\text{v}_{t,j} | s_t^\text{v}, a^\text{v}_{t, < j}))$
In the first term, lower entropy means higher semantic certainty. We encourage confident and decisive critiques. 
The second term ensures that the verifier only receives credit or punishment when the final answer is actually supported by the retrieved documents, mitigating bias from imperfect retrieval. 
The third term penalizes responses that filter correct answers, thereby promoting faithfulness.

\subparagraph{(3) Behavior Impact}
Most critically, we quantify how the verifier's feedback influences subsequent reasoning. Let $H^\text{r}_{a_t}$ denote the average token-level entropy of reasoner's action $a_t^\text{r}$. Over a dialogue of $n$ interaction steps, we analyze the entropy trend in the last three steps and classify it into one of the five patterns defined in Section~\ref{subsec:ent_pattern}. We then assign a score to each pattern: $\text{score}(p) = \{ \text{D} \to 1.0, \ \text{ID} \to 0.8, \ \text{F} \to 0.6, \ \text{DI} \to 0.4, \ \text{I} \to 0.2 \}$
Finally, the impact is then:
\begin{equation}
    \mathcal{A}_{\text{impact}} = \frac{1}{|\mathcal{A}^\text{r}_{\text{sub}}|} \sum^{|\mathcal{A}^\text{r}_{\text{sub}|}}_{j=1} \text{score}\left(f_e([H^\text{r}_{a_1}, ..., H^\text{r}_{a_n}])\right),
\end{equation}
where $\mathcal{A}^\text{r}_{\text{sub}}$ is the set of reasoner action subjected to monitoring.
This term incentivizes the verifier to provide feedback that steers the reasoner toward low-entropy and decisive reasoning. 

\begin{table*}[h]
\centering
\resizebox{0.93\textwidth}{!}{
\begin{tabular}{lcccccccccccccccc}
\toprule
\multicolumn{1}{c}{\multirow{3}{*}{Method}} & \multicolumn{6}{c}{\textbf{General QA}} & \multicolumn{8}{c}{\textbf{Multi-hop QA}} & \multicolumn{2}{c}{\multirow{2}{*}{\textbf{Average}}} \\
\cmidrule(lr){2-7} \cmidrule(lr){8-15}
& \multicolumn{2}{c}{\textbf{NQ}} & \multicolumn{2}{c}{\textbf{TriviaQA}} & \multicolumn{2}{c}{\textbf{PopQA}} & \multicolumn{2}{c}{\textbf{HotpotQA}} & \multicolumn{2}{c}{\textbf{2Wiki}} & \multicolumn{2}{c}{\textbf{Musique}} & \multicolumn{2}{c}{\textbf{Bamboogle}} \\
\cmidrule(lr){2-3} \cmidrule(lr){4-5} \cmidrule(lr){6-7} \cmidrule(lr){8-9} \cmidrule(lr){10-11} \cmidrule(lr){12-13} \cmidrule(lr){14-15} \cmidrule(lr){16-17}
\multicolumn{1}{c}{} & EM & F1 & EM & F1 & EM & F1 & EM & F1 & EM & F1 & EM & F1 & EM & F1 & EM & F1 \\
\midrule
\multicolumn{15}{l}{\textbf{\textit{Qwen2.5-3B Instruct}}} \\
CoT & 
2.5	& 8.7 &  
7.3	& 13.2 &  
7.4	& 13.7 &  
3.4	& 12.5 &  
2.1 & 12.4 &  
0.2	& 3.2 &
0.0 & 0.0 &
3.27 & 9.1 \\
RAG & 
33.7 & 39.6	& 
51.5 & 58.4 & 
36.5 & 44.2 & 
22.1 & 30.4 & 
21.6 & 29.9 & 
7.2 & 13.6 & 
60 & 12.3 & 
25.5 & 32.6  \\
Search-R1 &  
38.7 & 45.9 & 
55.2 & 66.7 &
40.0 & 50.2 &
31.4 & 38.3 & 
32.4 & 41.2 & 
10.2 & 20.8 &
18.9 & 25.6 &  
32.4 & 41.2 \\
\ \ - \textit{pass@2} &  
\underline{42.9} & - &  
\underline{64.5} & - &  
\underline{42.2} & - &  
\underline{34.1} & - & 
\underline{35.0} & - & 
\underline{12.8} & - &
\underline{25.6} & - &
\underline{36.7} & - \\
ReSearch &  
39.5 & \underline{46.7} &  
59.3 & \underline{67.8} &  
41.4 & 49.9 &  
32.9 & 38.6 &  
33.3 & 42.5 &  
11.4 & 21.0 &
24.7 & 31.2 &
34.6 & 42.5 \\
WebSeer &  
39.0 & 46.3 &  
58.5 & 67.1 &  
40.7 & \underline{50.6} &  
35.3 & \underline{44.4} &  
34.8 & \underline{43.6} &  
11.5 & \underline{21.7} &
24.3 & \underline{32.5} &
34.8 & \underline{43.7} \\
\algo &  
\textbf{43.7} & \textbf{53.2} &  
\textbf{65.6} & \textbf{73.6} &  
\textbf{46.6} & \textbf{54.1} &  
\textbf{36.0} & \textbf{44.9} &  
\textbf{35.8} & \textbf{44.6} &  
\textbf{14.5} & \textbf{23.7} &
\textbf{27.9} & \textbf{36.1} &
\textbf{38.6} & \textbf{47.2} \\
\midrule
\multicolumn{15}{l}{\textbf{\textit{Qwen2.5-7B Instruct}}} \\
CoT & 
3.5 & 11.0 & 
18.7 & 25.4 & 
8.8 & 14.1 & 
7.2 & 15.9 & 
6.7 & 15.2 & 
2.8 & 8.3 & 
13.8 & 19.6 & 
8.8 & 15.6\\
RAG & 
30.6 & 37.2 & 
58.0 & 64.2 & 
38.1 & 45.4 & 
26.4 & 34.0 & 
24.9 & 33.3 & 
7.4 & 14.7 & 
16.3 & 24.3 & 
28.8 & 36.2  \\
Search-R1 &  
40.6 & 48.0 & 
65.2 & \underline{73.7} & 
39.7 & 49.2 & 
38.4 & 45.6 & 
38.9 & 45.6 & 
14.7 & 21.4 &
37.1 & 44.0 &
39.2 & 46.8 \\
\ \ - \textit{pass@2} &  
41.5 & - & 
\underline{67.4} & - & 
\underline{50.6} & - & 
\underline{40.2} & - & 
40.6 & - & 
15.6 & - &
\underline{45.9} & - &
\underline{43.1} & - \\
ReSearch &  
\underline{41.6} & \underline{49.8} &  
64.0 & 71.6 &  
42.8 & \underline{50.8} &  
39.2 & \underline{46.0} &  
40.8 & 47.2 &  
\underline{15.8} & \underline{25.3} &
43.1 & \underline{49.7} &
41.0 & 48.6 \\
WebSeer &  
40.1 & 48.4 &  
65.5 & 73.2 &  
42.3 & 48.7 &  
38.6 & 45.6 &  
\textbf{41.7} & \textbf{51.3} &  
14.4 & 24.6 &
42.6 & 48.9 &
40.7 & \underline{48.7} \\
\algo &  
\textbf{45.1} & \textbf{53.7} &  
\textbf{68.2} & \textbf{75.5} &  
\textbf{50.8} & \textbf{56.6} &  
\textbf{45.5} & \textbf{54.2} &  
\underline{41.5} & \underline{50.6} &  
\textbf{17.7} & \textbf{27.3} &
\textbf{46.4} & \textbf{53.8} &
\textbf{45.0} & \textbf{53.1} \\
\midrule
\multicolumn{15}{l}{\textbf{\textit{Qwen3-8B}}} \\
CoT &  
2.7 & 10.7 & 
19.0 & 25.7 & 
8.5 & 14.7 & 
7.4 & 16.7 & 
7.8 & 16.9 & 
5.3 & 13.6 & 
17.4 & 26.3 & 
9.7 & 17.8  \\
RAG &  
34.1 & 42.7 & 
58.5 & 67.2 & 
40.5 & 47.2 & 
31.4 & 36.2 & 
30.7 & 35.4 & 
10.6 & 17.4 & 
23.1 & 31.7 & 
32.7 & 39.7  \\
Search-R1 &  
42.6 & 50.6 &  
67.5 & 75.1 &  
42.2 & 48.5 &  
41.8 & 47.3 &  
42.3 & 48.6 &  
16.3 & \underline{24.9} &
42.6 & 53.0 &
42.2 & 49.7 \\
\ \ - \textit{pass@2} &  
42.9 & - &  
\underline{70.5} & - &  
\underline{48.2} & - &  
45.4 & - &  
\underline{48.8} & - &  
\underline{19.6} & - &
\underline{48.9} & - &
\underline{46.3} & - \\
ReSearch &  
38.5 & 46.7 &  
62.3 & 70.8 &  
40.7 & 49.7 &  
38.1 & 47.4 &  
39.2 & 45.1 &  
15.2 & 23.1 &
40.6 & 49.7 &
39.2 & 47.5 \\
WebSeer &  
\underline{46.3} & \underline{52.8} &  
68.0 & \underline{75.6} &  
41.7 & \underline{50.4} &  
\underline{46.2} & \underline{53.4} &  
47.0 & \underline{54.8} &  
14.6 & 23.0 &
44.9 & \underline{55.6} &
44.1 & \underline{52.2} \\
\algo &  
\textbf{47.2} & \textbf{54.0} &  
\textbf{76.0} & \textbf{83.7} &  
\textbf{49.1} & \textbf{56.2} &  
\textbf{50.6} & \textbf{57.8} &  
\textbf{52.4} & \textbf{58.3} &  
\textbf{20.2} & \textbf{28.1} &
\textbf{53.4} & \textbf{64.7} &
\textbf{49.8} & \textbf{57.5} \\
\bottomrule
\end{tabular}
}
\caption{Performance comparison between \algo\ and baselines. Best and runner-up results are highlighted in \textbf{bold} and \underline{underline}.}
\label{tab:main_res}
\end{table*}

\subsubsection{Policy Optimization}
We optimize both agents using Group Relative Policy Optimization (GRPO), which normalizes advantages within a group of rollouts and incorporates a reference model for KL regularization. For each query $q$, we sample $G$ traces $\{\tau_i\}^G_{i=1}$ and calculate the outcome reward ${\{r_i^\alpha}\}^G_{i=1}$. The token-level advantage for $t$-th token in trace $i$ is first computed as:
$A^\alpha_{i,t} = \frac{r^\alpha_i - \text{mean}(r^\alpha_1, r^\alpha_2, ..., r^\alpha_G)}{\text{std}(r^\alpha_1, r^\alpha_2, ..., r^\alpha_G)}$
For the reasoner, the final advantage is $\hat{A}^\text{r}_{i,t} = A^\text{r}_{i,t}$. For the verifier, the process-aware advantage is added:
\begin{equation}
    \hat{A}^\text{v}_{i,t} = A^\text{v}_{i,t} + \mathbb{I}(y_{i, t} \in a_{t} \land a_t = \mathcal{A}_{sub}^\text{v}) \cdot A^\text{v}_{proc}
\end{equation}
ensuring $A_\text{proc}^\text{v}$ is only added to tokens belonging to the verifier's critique sections. Therefore, the policy model is optimized by maximizing:
\begin{align}
&\mathcal{J}^\alpha_{GRPO}(\theta) = \, 
\mathbb{E}_{\substack{q, \{ y_{i,t} \}_{i=1}^{G} \sim \\ \pi_{\text{old}}^\alpha( \cdot| x; \mathcal{R})}}
\Bigg[
\frac{1}{G} \sum_{i=1}^{G} \sum_{t=1}^{|y_i|}
\nonumber \\
&\min \Bigg( r^\alpha_i(\theta) \hat{A}^\alpha_{i,t}, \text{clip}(r^\alpha_i(\theta), 1-\epsilon,1+\epsilon)\hat{A}^\alpha_{i,t} \Bigg)
\nonumber \\
&- \beta \mathbb{D}_{KL} \left[ \pi^\alpha_{\theta} || \pi^\alpha_{\text{ref}} \right]
\Bigg],
\end{align}
where $r^\alpha_i(\theta) = \frac{\pi^\alpha_\theta(y_{i,t}|q)}{\pi^\alpha_\text{old}(y_{i,t}|q)}$ and $\pi_\text{ref}^\alpha$ is the reference model. Following common practices in this field, tokens not generated by the policy model $\pi^\alpha_{old}$ will be masked in the loss calculation.

\section{Experiments}
\label{sec:experiments}

\subsection{Setup}
\label{subsec:setup}

\begin{table}[h]
\centering
\resizebox{1.\columnwidth}{!}{
\begin{tabular}{lccccccc}
\toprule
Variants & \textbf{NQ} & \textbf{TQA} & \textbf{PQA} & \textbf{HQA} & \textbf{2Wiki} & \textbf{MSQ} & \textbf{BAM} \\
\midrule
\multicolumn{8}{l}{\textbf{\textit{Qwen2.5-3B Instruct}}} \\
\algo &  
53.2 &  
73.6 &  
54.1 &  
44.9 &  
44.6 &  
23.7 &
36.1 \\
\textit{w/o adv-out} &  
50.6 &  
69.6 &  
51.5 &  
43.2 &  
42.9 &  
22.4 &
35.5 \\
\textit{w/o proc-adv} &  
51.8 &  
69.0 &  
50.7 &  
43.6 &  
41.7 &  
20.6 &
33.4 \\
\midrule
\multicolumn{8}{l}{\textbf{\textit{Qwen2.5-7B Instruct}}} \\
\algo &  
53.7 &  
75.5 &  
56.6 &  
54.2 &  
50.6 &  
27.3 &
53.8 \\
\textit{w/o adv-out} &  
49.5 &  
73.6 &  
55.2 &  
50.4 &  
49.7 &  
25.3 &
52.6 \\
\textit{w/o proc-adv} &  
49.6 &  
72.5 &  
51.6 &  
45.2 &  
48.3 &  
21.9 &
47.0 \\
\midrule
\multicolumn{8}{l}{\textbf{\textit{Qwen3-8B}}} \\
\algo &  
54.0 &  
83.7 &  
56.2 &  
57.8 &  
58.3 &  
28.1 &
64.7 \\
\textit{w/o adv-out} &  
53.7 &  
83.4 &  
55.6 &  
57.3 &  
56.1 &  
27.5 &
63.2 \\
\textit{w/o proc-adv} &  
52.4 &  
82.5 &  
52.5 &  
54.6 &  
54.4 &  
23.9 &
61.8 \\
\bottomrule
\end{tabular}
}
\caption{Ablation studies on \algo. Datasets are abbreviated and correspond to Table~\ref{tab:main_res}, respectively.}
\label{tab:ablation}
\end{table}

\begin{figure}[h]
\centering
\includegraphics[width=0.95\columnwidth]{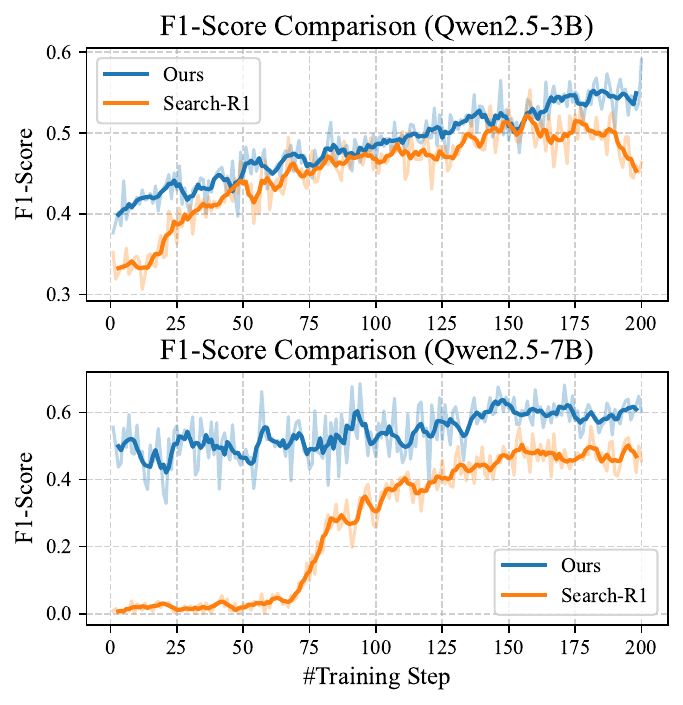}
\caption{F1-score comparison.}\label{fig:compare}
\end{figure}

\begin{figure}[h]
\centering
  \includegraphics[width=0.95\columnwidth]{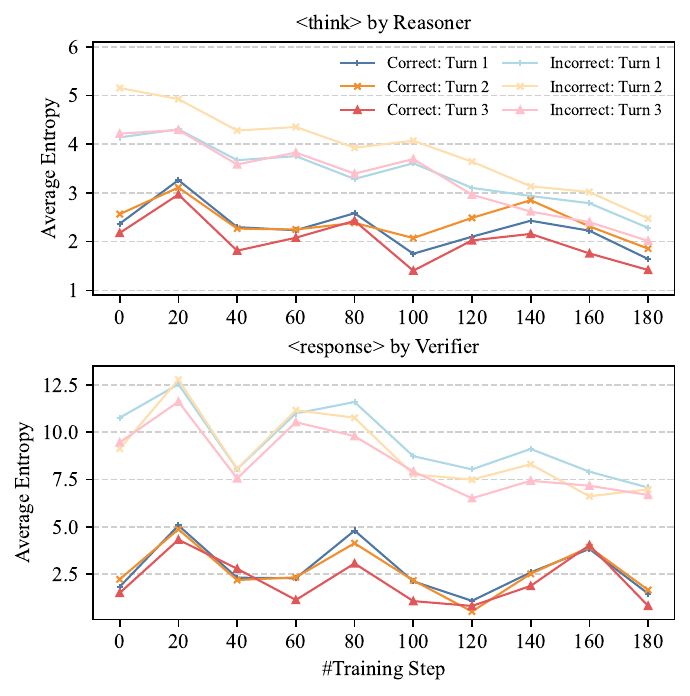}
  \caption {Entropy transition of agent actions in \algo.}
  \label{our_ent_pattern}
\end{figure}

\paragraph{Datasets \& Metrics}
We conduct evaluation on diverse QA benchmarks. Our method and all baselines are trained on NQ~\citep{nq_19_tom} and HotpotQA~\citep{hotpotqa_18_yang}. Following previous studies~\citep{deepre_25_zheng}, we randomly sample 512 examples from the development set of NQ, HotpotQA, TriviaQA~\citep{triviaqa_17_joshi}, 2WikiMultiHopQA~\citep{2wiki_20_ho}, PopQA~\citep{popqa_23_mallen}, and MuSiQue~\citep{musique_22_trivedi}, as well as all 125 samples from Bamboogle~\citep{bamboogle_23_press}. We adopt Exact Match (EM) and F1-score for comparison.

\paragraph{Baselines}
We compare our method against several baselines for reasoning and RAG in question answering, including CoT~\citep{cot_22_wei}, RAG~\citep{rag_20_lewis}, Search-R1~\citep{searchr1_25_jin}, ReSearch~\citep{research_25_chen}, and WebSeer~\citep{webseer_25_he}. Additionally, to fairly evaluate the efficacy of the adversarial yet cooperative dialogue in our multi-agent system, we introduce the \textbf{pass@2} metric for Search-R1.

\paragraph{Implementation}
For retrieval, all baselines adopt the same retriever and corpus setting as Search-R1. The retriever returns the top-3 documents. We select Qwen2.5-3B, -7B~\citep{qwen25}, and Qwen3-8B~\citep{qwen3} as backbone models. We optimize the policy model using the GRPO algorithm. For each prompt, we sample 5 trajectories with up to 5 interaction turns.

\subsection{Main Results}
Main results on general QA and multi-hop QA benchmarks across three backbone models are presented in Table~\ref{tab:main_res}. Overall, \algo\ consistently outperforms all baseline methods across varying model sizes and datasets. The average improvement over runner-up baseline is 11.1\% in EM and 7.6\% in F1-score on Qwen2.5-3B, 9.5\% in EM and 7.8\% in F1-score on Qwen2.5-7B, and 13.4\% in EM and 9.8\% in F1-score on Qwen3-8B. 
The performance gains of \algo\ remain consistent as the model size scales from 3B to 8B, suggesting that the proposed method is model-agnostic.

Remarkably, \algo\ with 3B backbone outperforms baselines models with 7B backbone on general QA benchmarks. This indicates that multi-perspective reasoning unleash the potential of compact backbones on relatively simple benchmarks. \algo\ also exhibits significant performance gains on several multi-hop QA benchmarks. For instance, the gains over runner-up on Musique is 26.1\%, 12.0\%, and 23.9\% in EM, respectively. Similarly, on HotpotQA, our method achieves the EM score of 0.455 and 0.506 on 7B and 8B. This shows that the multi-perspective reasoning architecture effectively solves complex multi-hop queries.

Our proposed method also frequently surpasses Search-R1 (\textbf{pass@2}). Take performance on NQ and HotpotQA with Qwen3-8B backbone for an example, \algo\ achieves 0.472 and 0.506 in EM, surpassing Search-R1 by 10\% and 11.5\%, respectively. This confirms that the superior performance of \algo\ is not the result of naive model scaling, but the adversarial yet cooperative dialogue and the multi-perspective optimization strategy. 

Figure~\ref{fig:compare} shows the F1-score of our method and Search-R1 throughout training. Our method consistently outperforms Search-R1. Unlike Search-R1 which suffers from the cold start problem, our methods shows strong performance during early training stage on the 7B model.

\subsection{Ablation Studies}
In this sub-section, we present the results of ablation experiments to evaluate the contribution of key components in \algo. We introduce 2 variants: (1) \algo\ without adversarial outcome rewards (\textit{w/o adv-out}) and (2) \algo\ without process-aware advantage (\textit{w/o proc-adv}). Results across three backbone models are shown in Table~\ref{tab:ablation}. 

The removal of the process-aware advantage leads to the most significant performance drop, particularly on multi-hop QA benchmarks. For instance, on Musique dataset with Qwen2.5-7B backbone, the F1 score drops from 27.3 to 21.9. This suggests that the proposed process-aware advantage is crucial for complex tasks requiring multi-step deduction.
The exclusion of the adversarial outcome reward also results in a consistent performance degradation, and the impact is smaller.

\subsection{Entropy Evolution}
We present the entropy transition of agent actions in multi-turn trajectories of \algo\ in Figure~\ref{our_ent_pattern}. In general, the action entropy of the third turn consistently achieving lower values than initial turns. The entropy of \texttt{response} by Verifier shows dramatic differences between correct and incorrect trajectories. These observations are consistent with the empirical studies regarding entropy pattern in Section~\ref{subsec:ent_pattern}.
The uncertainty within \texttt{think} by Reasoner gradually decreases as training progresses, indicating that the Reasoner is acquiring multi-turn reasoning capabilities. 

\section{Conclusion}
\label{sec:conclusion}
In this paper, we introduced \algo, a multi-perspective agentic RAG framework that decouples reasoning and verification into an adversarial yet co-evolving system. 
Further, 
we bridged the gap between outcome-oriented reward and process-aware guidance 
by proposing an adversarial outcome reward and a process-aware advantage that reward the verifier for evidence-grounded, and uncertainty reducing feedback. 
Results show that our methods consistently outperform existing baselines and frequently exceed the pass@2 results of competitors. 







\bibliography{custom}

\appendix
\section{Appendix}
\label{sec:apd}

\begin{figure*}[t]
\centering
  \includegraphics[width=1.0\linewidth]{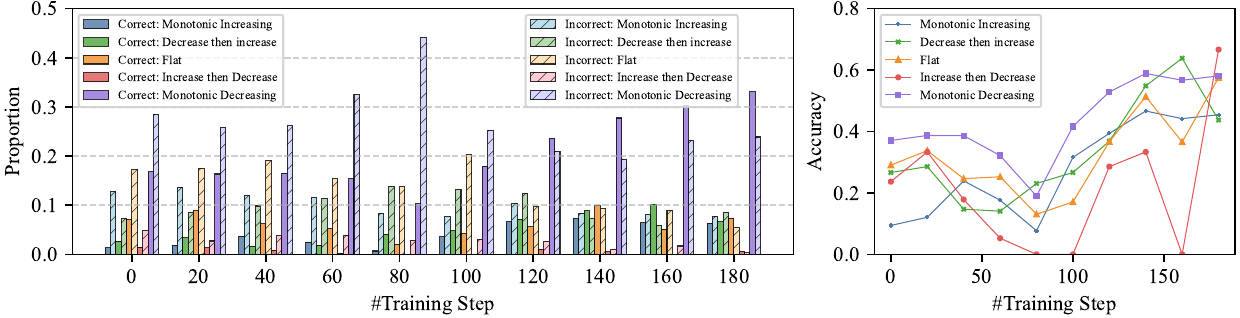}
  \caption {Statistical Analysis of Policy Entropy Pattern in Search-R1 trajectories. The y-axis of the \textbf{left subplots} denotes the proportion of trajectories exhibiting specific pattern relative to all multi-turn ($\ge 3$) samples. The y-axis of the \textbf{right subplots} represents the average accuracy of samples grouped by their pattens.}
  \label{fig:entpattern_acc-7b}
\end{figure*}

\subsection{Additional Preliminary Studies}
Due to the limited space, we present analysis of policy entropy pattern in Search-R1 trajectories on Qwen2.5-7B in this subsection. 
From Figure~\ref{fig:entpattern_acc-7b}, empirical studies on Qwen2.5-7B show similar pattern with studies on the 3B models. There is a positive correlation between correctness and decreasing entropy pattern. Similar to observations in Section~\ref{subsec:ent_pattern}, there exists a rise in the proportion of samples exhibiting an overall reduction in policy entropy. 
Quantitatively, for the Qwen2.5-7B backbone, the proportion rises from from 51.65\% in the early phase to 57.96\% in the late training phase.

\subsection{Implementation Details}
\label{subsec:implementation}
We use the 2018 Wikipedia dump~\citep{karpukhin2020densepassageretrievalopendomain} as the knowledge database and use E5~\citep{wang2024textembeddingsweaklysupervisedcontrastive} as the retriever model.
Our experiments are conducted on 8$\times$A100 GPUs, with full parameter optimization and gradient checkpointing. We build our method based on VeRL~\citep{sheng2024hybridflow} and use vLLM~\citep{kwon2023efficient} to accelerate agent rollouts. 

\subsection{Prompt Templates}
\label{apd:prompt}
\begin{prbox}[Prompt for the \textbf{Reasoner} \label{rea_prompt}]
To answer the given question, you will act as a reasoner working collaboratively with a retriever. Follow these steps carefully:

\vspace{2mm}
1. Reasoning Phase: When you receive a question, begin by reasoning about it inside \textcolor[HTML]{BC2929}{<think>} and \textcolor[HTML]{BC2929}{</think>}. This is where you analyze the problem and determine what you already know.

\vspace{2mm}
2. Identify Knowledge Gaps: If, during your reasoning, you realize that you lack some necessary information, you can request external knowledge by calling a search engine. To do this, write your query inside \textcolor[HTML]{2A5FBD}{<search>} and \textcolor[HTML]{2A5FBD}{</search>}.

\vspace{2mm}
3. Receive Search Results: After submitting your query, the verifier will process it and provide you with the top search results along with its opinion. This information will be enclosed between \textcolor[HTML]{6B9D32}{<feedback>} and \textcolor[HTML]{6B9D32}{</feedback>}.

\vspace{2mm}
4. Verification Phase: Every time you receive new information, you must first verify its relevance and usefulness. Conduct this verification inside \textcolor[HTML]{DFC43A}{<verify>} and \textcolor[HTML]{DFC43A}{</verify>}.

\vspace{2mm}
5. Update Reasoning: Based on the verified information, perform another round of reasoning inside \textcolor[HTML]{BC2929}{<think>} and \textcolor[HTML]{BC2929}{</think>}. Repeat steps 2–4 as many times as needed until you have enough information to answer the question.

\vspace{2mm}
6. Provide the Answer: Once you determine that no further external knowledge is required, provide your final answer directly inside \textcolor[HTML]{ED9136}{<answer>} and \textcolor[HTML]{ED9136}{</answer>}. Make sure to verify and think before answer the question. Keep your answer concise without additional explanations. For example: \textcolor[HTML]{ED9136}{<answer>} Beijing \textcolor[HTML]{ED9136}{</answer>}.

\vspace{2mm}
Always adhere strictly to the specified XML-like tags and respond only with the required elements.

\vspace{2mm}
Question: [QUESTION]

\label{tab:rea_prompt}
\end{prbox}

\begin{prbox}[Prompt for the \textbf{Verifier} \label{ver_prompt}]
As a verifier, your task is to collaborate with the reasoner to answer the given question. Follow these steps carefully:

\vspace{2mm}
1. Verification Process:

$\bullet$ The reasoner will provide its reasoning path, a retrieval query, and results from the search engine enclosed within \textcolor[HTML]{6B9D32}{<information>} ... \textcolor[HTML]{6B9D32}{</information>}.

$\bullet$ Perform a verification check inside \textcolor[HTML]{DFC43A}{<verify>} ... \textcolor[HTML]{DFC43A}{</verify>} to assess whether the query effectively contributes to answering the question.

\vspace{2mm}
2. Handling Effective Queries:

If the query is deemed appropriate:

$\bullet$ Choose the single most relevant document from the retrieved results and indicate it inside \textcolor[HTML]{875BC9}{<selected\_doc>} ... \textcolor[HTML]{875BC9}{</selected\_doc>} (e.g., \textcolor[HTML]{875BC9}{<selected\_doc>} Doc 1 \textcolor[HTML]{875BC9}{</selected\_doc>}).

$\bullet$ Synthesize the selected information and your own reasoning into a clear, concise reply inside \textcolor[HTML]{BC2929}{<response>}...\textcolor[HTML]{BC2929}{</response>}.

\vspace{2mm}
3. Handling Ineffective Queries:

If the query is judged ineffective, DIRECTLY Provide a justification for this assessment inside \textcolor[HTML]{BC2929}{<response>}...\textcolor[HTML]{BC2929}{</response>}.

\vspace{2mm}
4. Answer Verification:

If the reasoner provides an answer enclosed within \textcolor[HTML]{ED9136}{<answer>} and \textcolor[HTML]{ED9136}{</answer>}

$\bullet$ Verify the answer inside \textcolor[HTML]{DFC43A}{<verify>} ... \textcolor[HTML]{DFC43A}{</verify>} based on your judgment.

$\bullet$ Provide the final verified response inside \textcolor[HTML]{BA7127}{<final\_answer>}...\textcolor[HTML]{BA7127}{</final\_answer>}, ensuring it is concise and free of unnecessary details. For example: \textcolor[HTML]{BA7127}{<final\_answer>}Beijing\textcolor[HTML]{BA7127}{</final\_answer>}.

Always adhere strictly to the specified XML-like tags and respond only with the required elements.

\vspace{2mm}
Question: [QUESTION]

\label{tab:ver_prompt}
\end{prbox}

\begin{prbox}[Prompt for the \textbf{Final Predictor} \label{final_prompt}]
The rollout text of the reasoner and verifier is: [REASONER \& VERIFIER TRAJECTORY]

\vspace{2mm}
Answer the following question. Prior to this, both the reasoner and the verifier have conducted reasoning and verification regarding this question. You are required to provide the answer based on their respective reasoning processes. You should directly answer the question between without detailed illustrations.

\vspace{2mm}
Question: [QUESTION]

\label{tab:final_prompt}
\end{prbox}

\end{document}